\documentclass{article} 
\usepackage{nips15submit_e,times}
\usepackage{url}
\usepackage{caption}
\usepackage{graphicx}
\usepackage{amssymb,amsmath}
\usepackage{latexsym}
\usepackage[lined,boxed,commentsnumbered]{algorithm2e}
\newcommand{\comment}[1]{}
\newcommand{\mmp}[1]{}
\newcommand{\yaliwan}[1]{}

\newcommand{\pfm}{PFM}

\newcommand{\myemph}[1]{{\em #1}}
\newcommand{\mydef}[1]{{\em #1}}
\newcommand{\beq}{\begin{equation}}
\newcommand{\eeq}{\end{equation}}
\newcommand{\beqa}{\begin{eqnarray}}
\newcommand{\eeqa}{\end{eqnarray}}
\newcommand{\benum}{\begin{enumerate}}
\newcommand{\eenum}{\end{enumerate}}
\newtheorem{assumption}{Assumption}
\newtheorem{proposition}{Proposition}
\newtheorem{theorem}[proposition]{Theorem}
\newtheorem{lemma}{Lemma}

\newcounter{Examplecount}
\newcommand{\diag}{\operatorname{diag}}
\newcommand{\supp}{\operatorname{supp}}

\newcommand{\bfone}{{\mathbf 1}}
\newcommand{\rrr}{{\mathbb R}}
\newcommand{\bigOO}{{\cal O}}

\newcommand{\dtot}{d_{tot}}
\newcommand{\Rho}{\diag(\rho)}
\newcommand{\gmax}{g_{row}} 
\newcommand{\Vhat}{\hat{V}}
\newcommand{\Yhat}{\hat{Y}}
\newcommand{\Shat}{A}
\newcommand{\Phat}{\hat{P}}
\newcommand{\Lhat}{\hat{L}}

\newcommand{\Dhat}{\hat{D}}
\newcommand{\dhat}{\hat{d}}
\newcommand{\clust}{\mathcal{C}}

\newcommand{\gdet}{{\cal G}}
\newcommand{\hframe}{{\cal H}}
\newcommand{\nodes}{{\cal V}}

\newcommand{\epps}{\varepsilon}

\newcommand{\dmin}{d_{min}}
\newcommand{\dhatmin}{\dhat_{min}}
\newcommand{\epp}{\epsilon}
\newcommand{\Lamhat}{\hat{\Lambda}}
\newcommand{\lamhat}{\hat{\lambda}}

\newcommand{\delhat}{\hat{\delta}}
\newcommand{\err}{p_{err}}

\newenvironment{itemize*}{
\begin{itemize}
\setlength{\parskip}{0em}
\setlength{\topparskip}{0em}
}
{\end{itemize}}

\newenvironment{enumerate*}{
\begin{enumerate}
\setlength{\parskip}{0em}
\setlength{\topparskip}{0em}
}
{\end{enumerate}}

\newsavebox{\savestuff}
\newlength{\backitem}
\title{A class of network models recoverable by spectral clustering}
\author{
Yali Wan\thanks{ } \\
Department of Statistics\\
University of Washington\\
Seattle, WA 98195-4322, USA \\
\texttt{yaliwan@washington.edu} \\
\And
Marina Meil\u{a} \\
Department of Statistics \\
University of Washington \\
Seattle, WA 98195-4322, USA \\
\texttt{mmp@stat.washington.edu} \\
}

\nipsfinalcopy 

\begin{document}

\maketitle

\begin{abstract}
Finding communities in networks is a problem that remains difficult,
in spite of the amount of attention it has recently received.  The
Stochastic Block-Model (SBM) is a generative model for graphs with
``communities'' for which, because of its simplicity, the theoretical
understanding has advanced fast in recent years. In particular, there
have been various results showing that simple versions of spectral
clustering using the Normalized Laplacian of the graph can recover the
communities almost perfectly with high probability. Here we show that
essentially the same algorithm used for the SBM and for its extension
called Degree-Corrected SBM, works on a wider class of Block-Models,
which we call Preference Frame Models, with essentially the same
guarantees. \mmp{This is the maximal class on which current spectral
  techniques can be guaranteed to work.}  Moreover, the
parametrization we introduce clearly exhibits the free parameters
needed to specify this class of models, and results in bounds that
expose with more clarity the parameters that control the recovery
error in this model class.
\end{abstract}

\mmp{todo: what would happen if degrees much smaller? does correcting with tau or Gamma really improve something? -- i don't know if we can answer this question}

\section{Introduction}
\label{sec:intro}

\mmp{put some citations here.} 
There have been many recent advances in the recovery of communities in networks, under ``block-model'' assumptions \cite{rohe2011spectral, qin2013regularized, le2015}. In particular, advances in recovering communities by spectral clustering algorithms. These have been extended to models including node-specific propensities. In this paper, we argue that one can further expand the model class for which recovery by spectral clustering is possible, and describe a model that subsumes a number of existing models, which we call the \pfm. We show that under the \pfm~ model, the communities can be recovered with small error, w.h.p. Our results correspond to what \cite{chen2014statistical} termed the ``weak recovery'' regime, in which w.h.p. the fraction of nodes that are mislabeled is $o(1)$ when $n\rightarrow\infty$.

\section{The Preference Frame Model of graphs with communities}
\label{sec:model}
This model embodies the assumption that interactions at the community
level (which we will also call \myemph{macro} level) can be quantified
by meaningful parameters. This general assumption underlies the
$(p,q)$ and the related parameterizations of the SBM as well.  We define a
\mydef{preference frame} to be a graph with $K$ nodes, one for each
community, that encodes the connectivity pattern at the community
level by a (non-symmetric) {\em stochastic matrix} $R$. Formally, given
$[K]=\{1,\ldots K\}$, a $K\times K$ matrix $R$ (det$(R)\neq 0$) 
representing the transition matrix of a \myemph{reversible} Markov
chain on $[K]$, the weighted graph $\hframe=([K],R)$, with edge set $\supp R$ (edges correspond to entries in $R$ not being $0$)  is called a \mydef{$K$-preference frame}.
Requiring reversibility is equivalent to requiring that there is a set of \myemph{symmetric} weights on the edges from which $R$ can be derived (\cite{markovchains}). 
We note that without the reversibility assumption, we would be modeling
\myemph{directed} graphs, which we will leave for future work. We denote by $\rho$ the left principal eigenvector of $R$, satisfying $\rho^TR=\rho^T$. W.l.o.g. we can assume the eigenvalue 1 or $R$ has multiplicity 1\footnote{Otherwise the networks obtained would be disconnected.} and therefore we call $\rho$ the {\em stationary distribution} of $R$. 

We say that a deterministic weighted graph $\gdet=(\nodes, S)$ with
weight matrix $S$ (and edge set $\supp S$) \mydef{admits a $K$-preference frame $\hframe=([K],R)$} if
and only if there exists a partition $\clust$ of the nodes $\nodes$ into
$K$ clusters $\clust=\{C_1,\ldots\,C_k\}$ of sizes $n_1, \ldots, n_K$,
respectively, so that the Markov chain on $\nodes$ with transition matrix
$P$ determined by $S$ satisfies the linear constraints
\beq \label{eq:block-stoch}
\sum_{j\in C_m}\!P_{ij}\,=\,R_{lm}\quad
\text{for all $i\in C_l$},\;\text{and all cluster indices $l,m\in\{1,2,\ldots k\}$}.
 \eeq
 The matrix $P$ is obtained from $S$ by the standard row-normalization
 $P=D^{-1}S$ where $D=\diag\{d_{1:n}\},\,d_i=\sum_{i=1}^n S_{ij}$.
 
A random graph family over node set $\nodes$ \mydef{admits a $K$-preference frame $\hframe$}, and is called a {\em Preference Frame Model} (\pfm), if the edges $i,j,\,i<j$ are sampled independently from Bernoulli distributions with parameters $S_{ij}$. It is assumed that the edges obtained are undirected and that $S_{ij}\leq 1$ for all pairs $i\neq j$. We denote a realization from this process by $\Shat$. Furthermore, let $\dhat_i=\sum_{j\in\nodes}\Shat_{ij}$ and in general, throughout this paper, we will denote computable quantities derived from the observed $\Shat$ with the same letter as their model counterparts, decorated with the ``hat'' symbol. Thus, $\Dhat=\diag{\dhat_{1:n}},\,\Phat=\Dhat^{-1}\Shat$, and so on.

One question we will study is under what conditions the \pfm~model can be estimated from a given $\Shat$ by a standard spectral clustering algorithms. Evidently, the difficult part in this estimation problem is recovering the partition $\clust$. If this is obtained correctly, the remaining parameters are easily estimated in a Maximum Likelihood framework.

But another question we elucidate refers to the parametrization
itself. It is known that in the SBM and Degree Corrected-SBM (DC-SBM) \cite{qin2013regularized}, in spite of their
simplicity, there are dependencies between the community level ``intensive''
parameters and the graph level ``extensive''parameters, as we will
show below. In the parametrization of the \pfm~, we can explicitly
show which are the free parameters and which are the dependent ones.

Several network models in wide use admit a preference frame. For
example, the SBM$(B)$ model, which we briefly describe here. This
model has parameters the cluster sizes $(n_{1:K})$ and the {\em
connectivity matrix} $B\in[0,1]^{K\times K}$. For two nodes
$i,j\in \nodes$, the probability of an edge $(i,j)$ is $B_{kl}$ iff
$i\in C_k$ and $j\in C_l$. The matrix $B$ needs not be symmetric. When
$B_{kk}=p,B_{kl}=q$ for $k,l\in[K],\, k\neq l$, the model is denoted
SBM$(p,q)$. It is easy to verify that the SBM admits a preference frame. For instance, in the case of SBM$(p,q)$, we have
\[
d_i=p(n_l-1) + q(n-n_l)\equiv d_{C_l},\,\text{for}\,i\in C_l,\; 
\]
\[
R_{l,m} = \frac{q n_m }{d_{C_l}}\;\text{if}\,l\neq m,\,
R_{l,l} = \frac{p(n_l-1)}{d_{C_l}},\,
\text{for $l,m \in \{1,2,\ldots, k\}$.}
\]
In the above we have introduced the notation $d_{C_l}=\sum_{j\in C_l} d_i$. One particular realization of the \pfm~is the \mydef{Homogeneous
  $K$-Preference Frame} model (H\pfm). In a H\pfm, each node $i\in \nodes$ is characterized by a \mydef{weight, or propensity to form ties} $w_i$. For each pair of communities $l,m$ with
$l\leq m$ and for each $i\in C_l,j\in C_m$ we sample $\Shat_{ij}$ with
probability $S_{ij}$ given by  
\beq\label{eq:hpfm-sampling}
S_{ij}=\frac{R_{ml} w_i w_j}{\rho_{l}}.
\eeq
This formulation ensures detail balance in the
edge expectations, i.e. $S_{ij}=S_{ji}$. The H\pfm~is virtually equivalent to what is known as the ``degree model'' \cite{Jackson08} or ``DC-SBM'', up to a reparameterization\footnote{Here we follow the customary definition of this model, which does not enforce $S_{ii}=0$, even though this implies a non-zero probability of self-loops.}. Proposition \ref{prop:first} relates the node weights to the expected node degrees $d_i$. 
We note that the main result we prove in this paper uses independent sampling of edges only to prove the concentration of the laplacian matrix. The KPFM model can be easily extended to other graph models with dependent edges if one could prove concentration and eigenvalue separation. For example, when $R$ has rational entries, the subgraph induced by each block of $A$ can be represented by a random d-regular graph with a specified degree. \\
\begin{proposition}\label{prop:first}
In a H\pfm~$d_i=w_i\sum_{l=1}^KR_{kl}\frac{w_{C_l}}{\rho_l}$ whenever $i\in C_k$ and $k\in[K]$. 
\end{proposition}
Equivalent statements that the expected degrees in each cluster are proportional to the weights exist in \cite{C-O_Lanka09,rohe2011spectral} and they are instrumental in analyzing this model. This particular parametrization immediately implies in what case the degrees are globally proportional to the weights. This is, obviously, the situation when $w_{C_l}\propto \rho_l$ for all $l\in[K]$.

As we see, the node degrees in a H\pfm~are not directly determined by
the propensities $w_i$, but depend on those by a multiplicative
constant that varies with the cluster. This type of interaction
between parameters has been observed in practically all extensions of
the Stochastic Block-Model that we are aware of, making parameter
interpretation more difficult. Our following result
establishes what are the free parameters of the \pfm~and of their subclasses. As it will turn
out, these parameters and their interactions are easily interpretable.

\begin{proposition} \label{prop:nodedegree}
 Let \comment{ } $(n_1,\ldots n_K)$ be a partition of $n$ (assumed to represent the cluster sizes of $\clust=\{C_1,\ldots C_K\}$ a partition of node set $\nodes$), $R$ a non-singular $K\times K$ stochastic matrix, $\rho$ its left principal eigenvector, and $\pi_{C_1}\in [0,1]^{n_1},\ldots \pi_{C_K}\in [0,1]^{n_K}$ probability distributions over $C_{1:K}$.  Then, there exists a \pfm~ consistent with $\hframe=([K],R)$, with clustering $\clust$, and whose node degrees are given by 
\beq
d_i =\dtot \rho_k \pi_{C_k,i},
\eeq
whenever $i\in C_k$, where $\dtot=\sum_{i\in \nodes} d_i$ is a user parameter which is only restricted above by Assumption \ref{ass:s}.
\end{proposition}
The proof of this result is constructive, and can be found in the Supplement. 

The parametrization shows to what extent one can specify independently the degree distribution of a network model, and the connectivity parameters $R$. Moreover, it describes the pattern of connection of a node $i$ as a composition of a macro-level pattern, which gives the total probability of $i$ to form connections with a cluster $l$, and the micro-level distribution of connections between $i$ and the members of $C_l$. These
parameters are meaningful on their own and can be specified or
estimated separately, as they have no hidden dependence on each other
or on $n,K$.

The \pfm~enjoys a number of other interesting properties. As this
paper will show, almost all the properties that make SBM's popular and
easy to understand hold also for the much more flexible \pfm. In the
remainder of this paper we derive recovery guarantees for the \pfm.
As an additional goal, we will show that in the frame we set with the
\pfm, the recovery conditions become clearer, more interpretable, and
occasionally less restrictive than for other models.

\mmp{to keep these or not? Needs polishing\\
For example, in an SBM, the expected node degrees are the same for all
nodes in a community, and they grow proportionally to the size of the
community. Thus, no individual significance can be attached to the
parameters $p_{i}$ unless $n$ is fixed or node degrees are
unbounded. In the \cite{arora2012finding} node degrees are not bound to be
equal, however, to have a realistic model, one must postulate that a
node's propensity $p_i$ is lower when the node belongs to a larger
community. Thus, \myemph{local, individual} characteristics like the
propensity, are coupled with global network characteristics like the
community size, which again means that $p_i$ is not interpretable as a
truly local parameter.

In contrast, in our model, the parameters can be set individually and
independently\footnote{We do not claim that recovery is possible for
  any parameter settings, just that any parameter setting is possible
  and easily interpretable.}. Moreover, while here for simplicity we
will focus on graphs with fixed node set $V$, due to this independence
property, the \pfm~extends naturally to \mydef{mixture} models, where
one specifies a distribution of cluster sizes, and a degree
distribution for each $C_l$.}

Third, as already mentioned, the \pfm~includes many models that have
been found useful by previous authors. Yet, the \pfm~class is much
more flexible than those individual models, in the sense that it
allows other unexplored degrees of freedom (or, in other words,
achieves the same advantages as previously studied models with fewer
constraints on the data). Note that there is an infinite number of possible random
graphs $\gdet$ with the same parameters $(d_{1:n},n_{1:k},R)$ satisfying the constraints
\eqref{eq:block-stoch} and Proposition \ref{prop:nodedegree}, yet for
reliable community detection we do not need to estimate $S$ fully, but only to look at aggregate statistics like $\sum_{j\in C}A_{ij}$.

\section{Spectral clustering algorithm and main result}
\label{sec:alg-theorem}

\mmp{Hi Yali, you need to revise this as follows: (1) ``first
  eigenvectors'' --EVERYWHERE -- ``eigenvectors for what lambdas? this
  is important, fix it in the proofs as well.  (2) the table with
  notation belongs in the supplement.  (3) after main result: comment
  on the parameters: which depend only on preference frame, clusters,
  which are global, which depend on individual nodes.  (4) Modified
  Davis-Kahan -- who proved it? give citation (5) You have done quite
an impressive job writing and organizing so many results! Now you need
to go polish the writing just a bit, to make sure that terms are
defined before they are used, grammar is correct, etc.}

Now, we address the community recovery problem from a random graph
$(\nodes,A)$ sampled from the \pfm~defined as above. We make the
standard assumption that $K$ is known. Our analysis is based on a very
common spectral clustering algorithm used in \cite{MShi:nips00} and described also in  \cite{MShi:aistats01,von2007tutorial}. 
\begin{algorithm}[H] \label{algo:sc}
\SetKwData{Left}{left}\SetKwData{This}{this}\SetKwData{Up}{up}
\SetKwFunction{Union}{Union}\SetKwFunction{FindCompress}{FindCompress}
\SetKwInOut{Input}{Input}\SetKwInOut{Output}{Output}
\Input{Graph $(\nodes,A)$ with $|\nodes| = n$ and $A \in \{0, 1\}^{n \times n}$, number of clusters $K$}
\Output{Clustering $\clust$}
1. Compute $\Dhat = \diag(\dhat_1, \cdots, \dhat_n)$ and Laplacian
\beq
\Lhat =\Dhat^{-1/2}A\Dhat^{-1/2} \label{laplacian}
\eeq
2. Calculate the $K$ eigenvectors $\Yhat_1,\cdots, \Yhat_K$ associated with the $K$ eigenvalues $|\lamhat_1| \ge \cdots \ge |\lamhat_K|$ of $\Lhat$. Normalize the eigenvectors to unit length. We denote them as the first $K$ eigenvectors in the following text\;
3. Set $\Vhat_{i} = \Dhat^{-1/2}\Yhat_i$, $i=1,\cdots,K$. Form matrix $\Vhat=[\Vhat_1 \cdots \Vhat_K]$\;
4. Treating each row of $\Vhat$ as a point in $K$ dimensions, cluster them by the K-means algorithm to obtain the clustering $\hat{\clust}$.
\caption{Spectral Clustering}
\end{algorithm}
Note that the vectors $\Vhat$ are the first $K$ eigenvectors of $P$. The K-means algorithm is assumed to find the global optimum. For more details on good initializations for K-means in step 4 see \cite{ng2002spectral}.

We quantify the difference between $\hat{\clust}$ and the true clusterings $\clust$ by the mis-clustering rate $p_{err}$, which is defined as
\beq
p_{err} = 1 - \frac{1}{n} \max_{\phi:[K]\rightarrow[K]} \sum_k |C_{\phi(k)} \cap \hat{C}_k|.
\eeq
\begin{theorem}[Mis-clustering rate bound for H\pfm~ and \pfm] \label{thm: misclusterBound}
Let the $n\times n$ matrix $S$ admit a \pfm, and $w_{1:n}, R, \rho, P, A, d_{1:n}$ have the usual meaning, and let $\lambda_{1:n}$ be the eigenvalues of $P$, with $|\lambda_i|\geq|\lambda_{i+1}|$. Let  $\dmin = \min d_{1:n}$ be the minimum expected degree,  $\dhatmin=\min \hat{d}_i$, and $d_{max} = \max_{ij}nS_{ij}$. Let $\gamma \ge 1$, $\epsilon >0$ be arbitrary numbers. Assume:
\begin{assumption} \label{as:1}
$S$ admits a H\pfm~model and \eqref{eq:hpfm-sampling} holds. 
\end{assumption}
\begin{assumption}\label{ass:s}
$S_{ij} \leq 1$
\end{assumption}
\begin{assumption}\label{as:2}
$\dhatmin \ge \log n$ 
\end{assumption}
\begin{assumption}\label{as:3}
$\dmin \ge \log n$
\end{assumption}
\begin{assumption}\label{as:4}
$\exists \varkappa>0$, $d_{max}  \le \varkappa\log n$
\end{assumption}
\begin{assumption}\label{ass:gmax}
$\gmax > 0$, where  $\gmax$ is defined in Proposition \ref{prop:separation}.
\end{assumption}
\begin{assumption}\label{as:6}
 $\lambda_{1:K}$ are the eigenvalues of $R$, and $|\lambda_K| - |\lambda_{K+1}|\,=\,\sigma > 0$. 
\end{assumption}
We also assume that we run Algorithm 1 on $S$ and that K-means finds the optimal solution. Then, for $n$ sufficiently large, the following statements hold with probability at least 
$(1- 2\exp{\frac{-\epsilon^2}{2 + \epsilon/\sqrt{\log n}} })(1 - e^{-\gamma})$.\\
{\bf \pfm} Assumptions \ref{ass:s} - \ref{as:6} imply 
\beq
\err \le \frac{K\dtot}{n\dmin\gmax }\left[\frac{C_0\gamma^4}{\sigma^2\log n} + \frac{4\epsilon^2}{\dhat_{min}}\right]
\eeq
{\bf H\pfm} Assumptions \ref{as:1} - \ref{ass:gmax} imply
\beq
\err \le \frac{K\dtot}{n\dmin\gmax }\left[\frac{C_0\gamma^4}{\lambda_K^2\log n} + \frac{4\epsilon^2}{\dhat_{min}}\right]
\eeq
where $C_0$ is a constant.
\end{theorem}
Note that $\err$ decreases at least as $1/\log n$ when $\dhat_{min} = \dmin = \log n$. This is because $\dhat_{min}$ and $\dmin$ help with the concentration of $L$. Using Proposition \ref{prop:separation}, the distances between rows of $V$, i.e, the true centers of the k-means step, are lower bounded by $\gmax/\dtot$. After plugging in the assumptions for $\dmin, \dhat_{min}, d_{max}$, we obtain
\beq
\err \le \frac{K\varkappa(C_0\gamma^4 + 4\epsilon^2\sigma^2)}{\gmax\sigma^2\log n}.
\eeq
This shows that $\err$ decreases as $1 / \log n$. Of the remaining quantities, $\kappa$ controls the spread of the degrees  $d_{i}$, and $C_0$ depends on $\kappa$ and $\gamma$.   
 Notice that $\lambda_K$ and $\sigma$ are eigengaps in H\pfm~model and \pfm~model respectively and depend only on the preference frame, and likewise for $\gmax$. The eigengaps ensure the stability of principal spaces and the separation from the spurious eigenvalues, as shown in Proposition \ref{thm:bound-blocks}.
 
\subsection{Proof outline, techniques and main concepts}
\label{sec:eval-bound}

The proof of Theorem \ref{thm: misclusterBound} (given in the extended version of the paper) relies
on three steps, which are to be found in most results dealing with
spectral clustering. First, concentration bounds of the empirical Laplacian
$\Lhat$ w.r.t $L$ are obtained. There are various conditions under
which these can be obtained, and ours are most similar to the recent result
of \cite{le2015}. The other tools we use are Hoeffding bounds and
tools from linear algebra. Second, one needs to bound the perturbation of the
eigenvectors $Y$ as a function of the perturbation in $L$. This is based on
the pivotal results of Davis and Kahan, see e.g
\cite{qin2013regularized}. A crucial ingredient in these type of
theorems is the size of the eigengap between the invariant subspace
$Y$ and its orthogonal complement. This is a condition that is
model-dependent, and therefore we discuss the techniques we introduce
for solving this problem in the \pfm~ in the next subsection.

The third step is to bound the error of the K-means clustering
algorithm. This is done by a counting argument. The crux of this step
is to ensure the separation of the $K$ distinct rows of $V$. This,
again, is model dependent and we present our result below.  The
details and proof are in the Supplement. All proofs are for the \pfm; to specialize to the H\pfm, one replaces $\sigma$ with $|\lambda_K|$

\subsection{Cluster separation and bounding the spurious eigenvalues in the \pfm}
\label{sec:evals}

\begin{proposition}[Cluster separation]\label{prop:separation}
Let $V,\rho,d_{1:n}$ have the usual meaning and define the {\em cluster volume} $d_{C_k}=\sum_{i\in C_k}d_i$, and  $c_{max},c_{min}$  as $\max_k, min_k\frac{d_{C_k}}{n\rho_k}$.  Let $i,j\in\nodes$ be nodes belonging respectively to clusters $k,m$ with $k\neq m$. Then,
\beq
||V_{i:}-V_{j:}||^2\,\geq\,\frac{1}{\dtot} \left[
  \frac{1}{c_{max}}\left(\frac{1}{\rho_k}+\frac{1}{\rho_m}\right)-
  \frac{1}{\sqrt{\rho_k\rho_m}}\left(\frac{1}{c_{min}}-\frac{1}{c_{max}}\right) \right]\,=\,\frac{\gmax}{\dtot},
\eeq
where $\gmax =  \left[
  \frac{1}{c_{max}}\left(\frac{1}{\rho_k}+\frac{1}{\rho_m}\right)-
  \frac{1}{\sqrt{\rho_k\rho_m}}\left(\frac{1}{c_{min}}-\frac{1}{c_{max}}\right) \right]$.
Moreover, if the columns of $V$ are normalized to length 1, the above result holds by replacing $c_{max,min}$ with $\tilde{c}_{max,min}=\max,\min_k \frac{n_k}{n\rho_k}$. 
\end{proposition}
\mmp{notation gmax here}
In the square brackets, $c_{max,min}$ depend on the
cluster-level degree distribution, while all the other quantitities
depend only of the preference frame. Hence, this expression is
invariant with $n$, and as long as it is strictly positive, we have
that the cluster separation is $\Omega{(1/\dtot)}$.

The next theorem is crucial in proving that $L$ has a constant
eigengap. We express the eigengap of $P$ in terms of the preference
frame $\hframe$ and the mixing inside each of the clusters $C_k$. For
this, we resort to {\em generalized stochastic matrices},
i.e. rectangular positive matrices with equal row sums, and we relate
their properties to the mixing of Markov chains on bipartite graphs.

These tools are introduced here, for the sake of intuition, toghether
with the main spectral result, while the rest of the proofs are in the
Supplement.

Given $\clust$, for any vector $x \in \rrr^n$, we denote by $x_k,\,k=1,\ldots K$,
the block of $x$ indexed by elements of cluster $k$ of
$\clust$. Similarly, for any square matrix $A\in \rrr^{n\times n}$, we
denote by $A_{kl}=[A_{ij}]_{i\in k,j\in l}$ the block with rows
indexed by $i\in k$, and columns indexed by $j\in l$.

Denote by $\rho$, $\lambda_{1:K}$, $\nu^{1:K}\in \rrr^K$ respectively
the stationary distribution, eigenvalues\footnote{Here too,
  eigenvalues will always be ordered in decreasing order of their
  magnitudes, with positive values preceeding negatives one of the
  same magnitude. Consequently, for any stochastic matrix,
  $\lambda_1=1$ always}, and eigenvectors of $R$. 

We are interested in block stochastic matrices $P$ for which the
eigenvalues of $R$ are the principal eigenvalues. We call
$\lambda_{K+1}\ldots \lambda_{n}$ {\em spurious} eigenvalues. Theorem \ref{thm:bound-blocks} below is a sufficient condition that bounds $|\lambda_{K+1}|$
whenever each of the $K^2$ blocks of $P$ is "homogeneous" in a sense
that will be defined below.

When we consider the matrix $L=D^{-1/2}SD^{-1/2}$ partitioned according to $\clust$, it will be convenient to consider the off-diagonal blocks in pairs. This
is why the next result describes the properties of matrices consisting
of a pair of off-diagonal blocks.

\begin{proposition}[Eigenvalues for the off-diagonal blocks] \label{prop:singv}
Let $M$ be the square matrix
\beq \label{eq:M}
M=\left[\begin{array}{cc}0&B\\ 
		         A&0\\
			 \end{array}\right]
\eeq
where $A\in \rrr^{n_2\times n_1}$ and $B\in \rrr^{n_1\times n_2}$, and let $x=\left[\begin{array}{cc}x_1\\x_2\\\end{array}\right]$, $x_{1,2}\in{\mathbb C}^{n_{1,2}}$ be an eigenvector of $M$ with eigenvalue $\lambda$. Then
\beqa
Bx_2\;=\;\lambda x_1 & ABx_2\;=\;\lambda^2 x_2  \\
Ax_1\;=\;\lambda x_2 & BAx_1\;=\;\lambda^2 x_1  \\
M^2 &=\;\left[\begin{array}{cc}BA & 0\\ 0 & AB\\\end{array}\right]
\eeqa
Moreover, if $M$ is symmetric, i.e $B=A^T$, then $\lambda$ is a
singular value of $A$, $x$ is real, and $-\lambda$ is also an
eigenvalue of $M$ with eigenvector $[x_1^T\,-x_2^T]^T$. Assuming
$n_2\leq n_1$, and that $A$ is full rank, one can write $A=V\Lambda
U^T$ with $V\in \rrr^{n_2\times n_2}$, $U\in \rrr^{n_1\times n_2}$
orthogonal matrices, and $\Lambda$ a diagonal matrix of non-zero
singular values.  \end{proposition}

\begin{theorem}[Bounding the spurious eigenvalues of $L$]\label{thm:bound-blocks} Let $\clust,L,P,D,S,R,\rho$ be defined as above, and let $\lambda$ be an  eigenvalue of $P$. Assume that (1) $P$ is block-stochastic with respect to $\clust$; (2) $\lambda_{1:K}$ are the eigenvalues of $R$, and $|\lambda_K|>0$; (3) $\lambda$ is not an eigenvalue of $R$; (4) denote by $\lambda^{kl}_3$ ($\lambda^{kk}_2$) the third (second) largest in magnitude eigenvalue of block $M_{kl}$ ($L_{kk}$) and assume that $\frac{|\lambda^{kl}_3|}{\lambda_{max}(M_{kl})}\leq c<1$ ($\frac{|\lambda^{kk}_2|}{\lambda_{max}(L_{kk})}\leq c$). Then, the spurious eigenvalues of $P$ are bounded by $c$ times a constant that depends only on $R$.
\beq\label{eq:boundr} |\lambda|\leq
c\max_{k=1:K}\left(r_{kk}+\sum_{l\neq k}\sqrt{r_{kl}r_{lk}}\right)
\eeq
\end{theorem}
Remarks: The factor that multiplies $c$ can be further bounded denoteing $a=[\sqrt{r_{kl}}]^T_{l=1:K},b=[\sqrt{r_{lk}}]^T_{l=1:K}$
\beq
r_{kk}+\sum_{l\neq k}\sqrt{r_{kl}r_{lk}}\;=\;
a^Tb\;\leq||a||||b||\;=\;\sqrt{\sum_{l=1}^Kr_{kl}\sum_{l=1}^Kr_{lk}}
\;=\;\sqrt{\sum_{l=1}^Kr_{lk}}
\eeq
In other words, 
\beq
|\lambda|\leq \frac{c}{2}\max_{k=1:K}\sqrt{\sum_{l=1}^Kr_{lk}}
\eeq
The maximum column sum of a stochastic matrix is 1 if the matrix is doubly stochastic and larger than 1 otherwise, and can be as large as $\sqrt{K}$. However, one must remember that the interesting $R$ matrices have ``large'' eigenvalues. In particular we will be interested in  $\lambda_K>c$. It is expected that under these conditions, the factor depending on $R$ to be close to 1. 

The second remark is on the condition (3), that all blocks have small spurious eigenvalues. This condition is not merely a technical convenience. If a block had a large eigenvalue, near 1 or $-1$ (times its $\lambda_{max}$), then that block could itself be broken into two distinct clusters. In other words, the clustering $\clust$ would not accurately capture the cluster structure of the matrix $P$. Hence, condition (3) amounts to requiring that no other cluster structure is present, in other words that within each block, the Markov chain induced by $P$ {\em mixes well}.

\section{Related work}
\label{sec:related}

{\bf Previous results we used} The Laplacian concentration results
use a technique introduced recently by \cite{le2015}, and some of the basic
matrix theoretic results are based on \cite{MShi:aistats01} which
studied the $P$ and $L$ matrix in the context of spectral clustering.
As any of the many works we cite, we are indebted to the pioneering
work on the perturbation of invariant subspaces of Davis and Kahan
\cite{qin2013regularized, rohe2011spectral, stewart1990matrix}.

\subsection{Previous related models}
The configuration model for regular random
graphs \cite{Bollobas_RG01,McKay_reg90} and for graphs with general
fixed degrees \cite{McKay_degs85, McKay_gen_degs91} is very well
known.  It can be shown by a simple calculation that the configuration
model also admits a $K$-preference frame.  In the particular case when
the diagonal of the $R$ matrix is $0$ and the connections between
clusters are given by a bipartite configuration model with fixed
degrees, $K$-preference frames have been studied by
\cite{Newman_equitable14} under the name ``equitable graphs''; the
object there was to provide a way to calculate the spectrum of the
graph.

Since the \pfm~ is itself an extension of the SBM, many other
extensions of the latter will bear resemblance to \pfm. Here we
review only a subset of these, a series of strong relatively recent
advances, which exploit the spectral properties of the SBM and extend
this to handle a large range of degree distributions 
\cite{C-O_Lanka09,rohe2011spectral,chaudhuriChungAtsias:12}. The \pfm~ includes each of these models as a subclass\footnote{In
  particular, the models proposed in
  \cite{C-O_Lanka09,rohe2011spectral,chaudhuriChungAtsias:12} are
  variations of the DC-SBM and thus forms of the homogeneous \pfm.}.

In \cite{C-O_Lanka09} the authors study a model that coincides
(up to some multiplicative constants) with the H\pfm. The paper
introduces an elegant algorithm that achieves partial recovery or
better, which is based on the spectral properties of a random
Laplacian-like matrix, and does not require knowledge of the partition
size $K$.  \mmp{CHECK WHAT IS NEW NOW We, instead of requiring a minimum average degree\footnote{This is what assumption C4 in
  \cite{C-O_Lanka09} essentially implies.} and minimum cluster
size, we will introduce a \myemph{saliency} condition similar to the
Informal Theorem in Section 2, that balances community size with
average degree, essentially requiring that average degree is large
only in communities that are small, instead of uniformly over all
nodes. Such a condition is also more conducive results in the sparse regime. }

The \pfm~also coincides with the model of \cite{arora2012finding} and
\cite{Jackson08} called the \myemph{expected degree model} w.r.t the distribution of \myemph{intra-cluster} edges, but not w.r.t the ambient edges, so the H\pfm~ is a subclass of this model. 

{\bf A different approach to recovery} The papers
\cite{chaudhuriChungAtsias:12,qin2013regularized,le2015} 
propose regularizing the normalized Laplacian with respect to
the influence of low degrees, by adding the scaled unit matrix $\tau
I$ to the incidence matrix $\Shat$, and thereby they achieve recovery
for much more imbalanced degree distributions than us. Currently, we
do not see an application of this interesting technique to the \pfm,
as the diagonal regularization destroys the separation of the
intracluster and intercluster transitions, which guarantee the
clustering property of the eigenvectors. Therefore, currently we
cannot break the $n\log n$ limit into the ultra-sparse regime,
although we recognize that this is an important current direction of research.

Recovery results like ours can be easily extended to weighted,
non-random graphs, and in this sense they are relevant to the spectral
clustering of these graphs, when they are assumed to be noisy versions
of a $\gdet$ that admits a \pfm.

\subsection{An empirical comparison of the recovery conditions}
\label{sec:empirical}

As obtaining general results in comparing the various recovery
conditions in the literature would be a tedious task, here we
undertake to do a numerical comparison. While the conclusions drawn
from this are not universal, they illustrate well the stringency of
various conditions, as well as the gap between theory and actual
recovery. For this, we construct H\pfm~ models, and verify numerically
if they satisfy the various conditions. We have also clustered random
graphs sampled from this model, with good results (shown in the
Supplement).

\mmp{In this session, we will use simulations to show that our theorem \ref{thm: misclusterBound} fits into the data generated from H\pfm~model while the assumptions of the other existing methods cannot be satisfied with this model.}

We generate $S$ from the H\pfm~model with $K = 5$, $n = 5000$. Each $w_i$ is uniformly generated from $(0.5, 1)$. $n_{1:K} = (500 ,1000 ,1500 ,1000 ,1000)$, $\gmax > 0$, $\lambda_{1:K} =(1, 0.8, 0.6, 0.4, 0.2)$. The matrix $R$ is given below; note its last row in which $r_{55}<\sum_{l=1}^4r_{5l}$. 
\\ \yaliwan{Should probably add assumptions on $\gmax$ here as well}
\begin{small}
\beq
R = \begin{pmatrix}
.80 & .07 & .02 & .02 & .09\\
.04 & .52 & .24 & .12 & .08\\
.01 & .20 & .65 & .15 & .00\\
.01 & .08 & .12 & .70 & .08\\
.13 & .21 & .02 & .32 & .33
\end{pmatrix}
\quad
\rho = (.25, .44, .54, .65, .17).
\eeq
\end{small} 
The conditions we are verifying include besides ours, those obtained by \cite{qin2013regularized}, \cite{rohe2011spectral}, \cite{balcanBorgsBravermanChayesTeng:12} and \cite{chaudhuriChungAtsias:12}; since the original $S$ is a perfect case for spectral clustering of weighted graphs, we also verify the theoretical recovery conditions for spectral clustering in \cite{balakrishnan2011noise} and \cite{ng2002spectral}. 

{\bf Our result Theorem \ref{thm: misclusterBound}} We have $\dmin = 77.4$, $\dhatmin = 63$, both bigger than $\log n = 8.52$. Therefore assumptions $(a)$ and $(b)$ hold; $d = 2500$,  so $(c)$ also holds, $\gmax = 1.82>0$.
After running the algorithm, the mis-clustering is rate $r=0.0008$, which  satisfies the theoretical bound. In conclusion, the dataset fits into both the assumptions and conclusion of Theorem \ref {thm: misclusterBound}.\\
{\bf Qin and Rohe}\cite{qin2013regularized} This paper has an assumption on the lower bound on $\lambda_K$, that is $\frac{1}{8\sqrt{3}}\lambda_K \ge \sqrt{\frac{K(ln(K/\epp)}{\dmin}}$, so that the concentration bound holds with probability $(1-\epp)$. We set $\epp = 0.1$ and obtain $ \lambda_K \ge 12.3$, which is impossible to hold since $\lambda_K$ is upper bounded by $1$\footnote{To make $\lambda \le 1$ possible, one needs $\dmin \ge 11718$.\mmp{what do you mean lambda$\ge 1$??}}.\\
{\bf Rohe, Chatterjee, Yu}\cite{rohe2011spectral} Here, one defines $\tau_n = \frac{\dmin}{n}$, and requires $\tau_n^2\log n > 2$ to ensure the concentration of $L$. To meet this assumption, with $n = 5000$, $\dmin \ge 2422$. While in our case $\dmin = 77.4$. The assumption requires a very dense graph and is not satisfied in this dataset.\\
{\bf Balcan, Borgs Braverman, Chayes}\cite{balcanBorgsBravermanChayesTeng:12}Their theorem is based on self-determined community structure. It requires all the nodes to be more connected within their own cluster. However, in our graph, $1296$ out of $5000$ nodes have more connections to outside nodes than to nodes in their own cluster.\\
\comment{
{\bf Balakrishnan, Xu, Krishnamurthy, Singh}\cite{balakrishnan2011noise} This paper assumes that $S$ is  the sum of an ideal matrix and a perturbation matrix. The conditions for $K$-way spectral clustering, require bounding the scale factor $\sigma$ of the perturbation matrix and ensure $\sigma = o(\frac{\beta_0}{K}(\frac{n }{Klogn})^{1/4})$. In our example, $\sigma = 0.23$, $\frac{\beta_0}{K}(\frac{n }{Klogn})^{1/4} = 0.32$. Therefore the assumption does not hold. \\}
{\bf Ng, Jordan, Weiss}\cite{ng2002spectral}  require $\lambda_2 < 1-\delta$, where $\delta > (2+2\sqrt{2})\epp$, $\epp = \sqrt{K(K-1)\epp_1 + K\epp_2^2}$, $\epp_1 \ge \max_{i_1, i_2 \in\{1,\cdots, K\}}\sum_{j\in C_{i_1}}\sum_{k \in C_{i_2}} \frac{A_{jk}^2}{\dhat_j\dhat_k}$, $\epp_2 \ge \max_{i \in \{1, \cdots, K\}} \frac{\sum_{k:k\in S_i}}{\dhat_j} (\sum_{k,l\in S_i} \frac{A_{kl}^2}{\dhat_k\dhat_l})^{1/2}$. On the given data, we find that $\epp \ge 36.69$, and $\delta \ge 125.28$, which is impossible to hold since $\delta$ needs to be smaller than $1$.\\
{\bf Chaudhuri, Chung, Tsiatas}\cite{chaudhuriChungAtsias:12} The recovery theorem of this paper requires $d_i \ge \frac{128}{9}\ln(6n / \delta)$, so that when all the assumptions hold, it recovers the clustering correctly with probability at least $1 - 6\delta$. We set $\delta = 0.01$, and obtain that $d_i = 77.40$, $\frac{128}{9}\ln(6n / \delta) = 212.11$. Therefore the assumption fails as well.\\
For our method, the hardest condition to satisfy, and the most
different from the others, was Assumption \ref{ass:gmax}.  We repeated
this experiment with the other weights distributions for which this
Assumption fails. The assumptions in the related papers continued to
be violated. In [Qin and Rohe], we obtain $\lambda_K \ge 17.32$. In [Rohe, Chatterjee, Yu], we still needs $\dmin \ge 2422$. In [Balcan, Borgs Braverman, Chayes], we get $1609$ points more connected to the outside nodes of its cluster. In [Balakrishnan, Xu, Krishnamurthy, Singh], we get $\sigma = 0.172$ and needs to satisfy $\sigma = o(0.3292)$.  In [Ng, Jordan, Weiss], we obtain $\delta \ge 175.35$. Therefore, the assumptions in these papers are all violated as well.

\section{Conclusion}
\label{sec:discussion}
In this paper, we have introduced the preference frame model, which is more flexible and subsumes many current models including SBM and DC-SBM. It produces state-of-the art recovery rates comparable to existing models. To accomplish this, we used a parametrization that is clearer and more intuitive. The theoretical results are based on the new geometric techniques which control the eigengaps of the matrices with piecewise constant eigenvectors. 

We note that the main result theorem 3 uses independent sampling of edges only to prove the concentration of the laplacian matrix. The KPFM model can be easily extended to other graph models with dependent edges if one could prove concentration and eigenvalue separation. For example, when $R$ has rational entries, the subgraph induced by each block of $A$ can be represented by a random d-regular graph with a specified degree. 



\section{Matrix theoretical results}
This proposition below collected various basic facts moved elsewhere in the references\cite{MShi:nips00,von2007tutorial}.
\begin{proposition}[\cite{MShi:nips00,von2007tutorial}] \label{prop:1}
\benum
\item The matrices $P$ and $L$ have the same eigenvalues, denoted $\lambda_{1:n}$ (slightly abusively).
\item Every eigenvalue of $R$ is also an eigenvalue of $P$. 
\item If $v$ is an eigenvector of $P$ with eigenvalue $\lambda$, then $u=D^{1/2}v$ is is an eigenvector of $L$ with the same eigenvalue.
\item In particular, $P\bfone=\bfone$ is the Frobenius vector of $P$, and therefore $Ls=s$, with $s_i=\sqrt{d_i}$, for $i=1,\ldots n$, is the Frobenius eigenvector of $L$. 
\item The stationary distribution of $P$ is $\pi$, with $\pi_i\propto d_i$, $i=1,\ldots n$. 
\comment{\item The probability of cluster $k$ under the stationary distribution is equal to $\rho_k$, and therefore $\rho_k\propto d_k$, with $d_k=\sum_{i\in k}d_i$ the volume of cluster $k$. CHECK THIS}
\item $R$ is diagonalizable, implying that it has $K$ independent eigenvectors. (This follows from the fact that the Markov chain defined by $P$ is reversible, implying that $R$ also defines a reversible Markov chain.)
\item Let $\lambda$ be an eigenvalue of $R$, $\nu$ its eigenvector, and $v$ the eigenvector of $P$ corresponding to $\lambda$.  Then $v_i=\nu_l$ whenever $i\in l$.  In other words, $P$ has $K$  eigenvectors that are "telescoped" versions of the eigenvectors of $R$. 
\eenum
\end{proposition}

Let $S,w_{1:n},K,R,\rho,d_{1:n},P$, etc have the usual meaning. Let
$B=\Rho^{-1/2}R\Rho^{-1/2}$. Denote \beqa X\in\rrr^{K\times K}&&
\text{the eigenvector matrix of $B$, orthonormal}\\ U\in\rrr^{K\times
  K}&& \text{the eigenvector matrix of $R$,
  with}\;\Rho^{1/2}U\,=\,X\\ Y\in\rrr^{n\times K}&& \text{principal
  eigenvectors of $L$, orthonormal},\;
Y_{il}\,\propto\,\frac{\sqrt{d_i}X_{kl}}{\sqrt{\rho_k}}\;\text{if
}i\in C_k\\ y_{1:K}&& \text{normalization constants for the columns of
  $Y$,}\\ V\in\rrr^{n\times K}&& \text{principal eigenvectors of
  $P$, with},\; V_{ik}=\frac{1}{\sqrt{d_i}}Y_{ik} \eeqa Denote also
$\dtot=\sum_{i=1}^nd_i,\,d_k=\sum_{i\in
  C_k}d_i,\pi_k=\frac{d_k}{\dtot}$, and
$\max_k,\min_k\frac{\pi_k}{\rho_k}=c_{max,min}$.  

{\bf Proof of Proposition 2 from the paper} We construct a
distribution $\pi$ over $\nodes$ by $\pi'=[\pi_1' \ldots\pi_K']$ with
$\pi_k\in [0,1]^{n_k}$ the elements of $\pi$ indexed by cluster
$C_k$. Let $\pi_k=\pi_{C_k}\rho_k$ for all $k=1,\ldots K$. 

We will verify that $\pi$ is the stationary distribution of $P$. Fix $i\in C_l$. We calculate $\pi'P$ at node $i$. 
\beqa
(\pi'P)_i&=&\sum_{k=1}^K \pi_k'P_{kl,:i}\;=\;
     \sum_{k=1}^K \rho_k\pi_{C_k}'\tilde{P}_{kl,:i}r_{kl}\;=\;
     \sum_{k=1}^K \rho_k\pi_{C_l,i}r_{kl}\;=\;
     \pi_{C_l,i}\sum_{k=1}^K \rho_kr_{kl}\;=\;
     \pi_{C_l,i}\rho_l\;=\;\pi_i
\eeqa
Above, we slightly abused notation by using $i$ both as an index in $\nodes$ and in $C_l$. 

It remains to show that $\pi\bfone=1$ which is straightforward and
left to the reader. Now, from Proposition \ref{prop:1} $\pi_i\propto
d_i$ for $i=\in\nodes$; the normalization constant for the r.h.s. of
this expression is $\sum_{i\in \nodes}d_i=\dtot$.

{\bf Proof of Proposition 4 from the paper}
Define   \beq \label{eq:separation}
y_l^2\,=\,\sum_i\frac{d_iX_{kl}^2}{\rho_k}\,=\,\sum_{k=1}^K\frac{d_kX_{kl}^2}{\rho_k}.
\eeq Hence, \beq c_{min}\dtot\,\leq \,y_l^2\,\leq\,
c_{max}\dtot\sum_{k}X_{kl}^2=c_{max}\dtot.  \eeq.  Now, we can derive
bounds on $||V_{i:}||^2$, the length of row $i$ of $V$.  \beq
||V_{i:}||^2\,=\,\sum_{l=1}^K\frac{1}{y_l^2}\frac{X_{kl}^2}{\rho_k}
\,=\,\frac{1}{\rho_k}\sum_{l=1}^K\frac{1}{y_l^2}X_{kl}^2
\,\geq\,\frac{1}{\rho_k}\frac{1}{\dtot c_{max}} \quad\text{when }i\in
C_k.  \eeq 
We now need to bound the cross-terms $V_{i:}^TV_{j:}$. Denote by $l^+=\{l\in [K],\, X_{kl}X_{ml}\geq 0\}$ and $l^-=[K]\setminus l^+$. Then, we have
\beqa
\sum_{l=1}^K\frac{1}{y_l^2}X_{kl}X_{ml}
&=&\sum_{l^+}\frac{1}{y_l^2}X_{kl}X_{ml}-\sum_{l^-}\frac{1}{y_l^2}|X_{kl}X_{ml}|\\
&\leq&\frac{1}{c_{min}\dtot}X^+-\frac{1}{c_{max}\dtot}X^+
\;=\;\left(\frac{1}{c_{min}\dtot}-\frac{1}{c_{max}\dtot}\right)X^+
\eeqa
where $X^+=\sum_{l^+}X_{kl}X_{ml}=\sum_{l^-}|X_{kl}X_{ml}|=\frac{1}{2}\sum_{l=1}^K|X_{kl}X_{ml}|\leq \frac{1}{2}$, because $X_{m:}\perp X_{k:}$. 

Now, putting these together we obtain the desired result. 
\beqa
||V_{i:}-V_{j:}||^2&=&||V_{i:}||^2+||V_{j:}||^2-2V_{i:}^TV_{j:}\\ &\geq&
\frac{1}{\dtot c_{max}}\left(\frac{1}{\rho_k}+\frac{1}{\rho_m}\right)-
\frac{2}{\sqrt{\rho_k\rho_m}}\frac{1}{2}\left(\frac{1}{c_{min}\dtot}-\frac{1}{c_{max}\dtot}\right)\\
 &=&
\frac{1}{\dtot} \left[
  \frac{1}{c_{max}}\left(\frac{1}{\rho_k}+\frac{1}{\rho_m}\right)-
  \frac{1}{\sqrt{\rho_k\rho_m}}\left(\frac{1}{c_{min}}-\frac{1}{c_{max}}\right) \right]
\eeqa 
In the case when the columns of $V$ are normalized, we have that 
\beq
V_{il}^{norm}\;=\;\frac{1}{v_l}\frac{X_{kl}}{\sqrt{\rho_k}}
\quad\text{whenever $i\in k$}
\eeq
and $v_l^2=\sum_{k=1}^Kn_k\frac{X_{kl}^2}{\rho_k}$\yaliwan{similarly to \eqref{eq:yl}}. Hence $\tilde{c}_{min}=\min_k\frac{n_k}{n\rho_k} \leq v_l^2\leq \tilde{c}_{max}=\max_k \frac{n_k}{n\rho_k}$. From this the second result follows.

\mmp{comments to move elsewhere}
In the square brackets, $c_{max,min}$ depend on $\pi_{1:k}$ the
degree distribution, while all the other quantities depend only of
the preference frame. Hence, this expression is invariant with $n$,
and as long as it is strictly positive, we have that the cluster separation is $\Theta(n^{-2})$. 

In particular, when all $\rho_k$ are equal, $c_{max}\leq 2c_{min}$ suffices. These bounds are not tight since $2X^+$ is never 1.

We firstly establish the main facts needed for the proof. These are properties of the eigenvalues for the off-diagonal blocks of $L$. 
\begin{proposition}[Maximum eigenvalue of a block of $L$]\label{prop:maxL} 

1) $L_{kk}s_k=r_{kk}s_k$ and therefore $\lambda_{max}(L_{kk})=r_{kk}$ for $k=1,\ldots K$.
2) $L_{kl}s_l=r_{kl}s_k$ and $\lambda_{max}(M_{kl})=\sqrt{r_{kl}r_{lk}}$ for all $k,l=1,\ldots K$, with $k\neq l$ and 
\beq
M_{kl}\;=\;\left[\begin{array}{cc}0& L_{kl}\\ L_{lk} & 0\\\end{array}\right]
\eeq
\end{proposition}

{\bf Proof} For part 1, note that
$L_{kk}=\operatorname{diag}(s_k)^{-1}S_{kk}\operatorname{diag}(s_k)^{-1}$
and that $\frac{1}{r_{kk}}P_{kk}$ is a stochastic matrix. Then, by
Proposition \ref{prop:1}
$\lambda_{max}(L_{kk})=\lambda_{max}(P_{kk})=r_{kk}$, and the
corresponding eigenvector of $L_{kk}$ is $s_k$.

For part 2, we show that the vector $x=[s'_k/\sqrt{\rho_k}\;s'_l/\sqrt{\rho_l}]'$ is an eigenvector of $M_{kl}$. 
\beq
M_{kl}x\;=\;
\left[\begin{array}{l} L_{kl}s_l/\sqrt{\rho_l}\\
                                 L_{lk}s_k/\sqrt{\rho_k}\\
              \end{array}\right]
\;=\;
\left[\begin{array}{l} r_{kl}s_k/\sqrt{\rho_l}\\
                       r_{lk}s_l/\sqrt{\rho_k}\\
              \end{array}\right]
\;=\;
\left[\begin{array}{l} \sqrt{r_{kl}r_{lk}}s_k\sqrt{\frac{r_{kl}}{r_{lk}\rho_l}}\\
                       \sqrt{r_{kl}r_{lk}}s_l\sqrt{\frac{r_{lk}}{r_{kl}\rho_k}}\\
              \end{array}\right]
\;=\;
\sqrt{r_{kl}r_{lk}}\left[\begin{array}{l} s_k/\sqrt{\rho_k}\\
                         s_l/\sqrt{\rho_l}\\
              \end{array}\right]
\eeq
The last equality holds because
\beq
\frac{r_{kl}}{r_{lk}\rho_l}\;=\;\frac{r_{kl}\rho_k}{r_{lk}\rho_l}\frac{1}{\rho_k}\;=\;\frac{1}{\rho_k}
\eeq
using the detailed balance of the reversible Markov chain defined by $R$. Since the eigenvector $x$ is positive, it must correspond to the largest eigenvalue of $M_{kl}$. 

\begin{proposition}\label{prop:perp}
Let $x$ be a {\em spurious eigenvector} of $L$, associated to spurious eigenvalue $\lambda$, and denote by $s_k,x_k\in \rrr^{n_k}$ the restrictions of $s,x$ to cluster $k$. Then $x_k\perp s_k$ for all $k=1,\ldots K$. 
\end{proposition}

{\bf Proof} Let $u$ be a principal eigenvector of $L$, and $u_k$ its
$k$-th block. From Proposition \ref{prop:1} it follows that
$u_k=s_k\nu_k$, where $\nu$ is an eigenvector of $R$. Because $L$ is symmetric, we know that $x\perp u$, which can be written equivalently as
\beq \label{eq:perp1}
\sum_{k=1}^K x_k's_k \nu_k\;=\;0
\eeq
Let $\xi_k=x_k's_k$. If we write \eqref{eq:perp1} for all $K$ eigenvectors of $R$, we obtain the linear system
\beq \label{eq:perp}
[\nu_1\;\nu_2\;\ldots \nu_K]\xi_k\;=\;0
\eeq
Since the matrix $[\nu_1\;\nu_2\;\ldots \nu_K]$ is non-singular, the system admits only the trivial solution $\xi_k=0$. \hfill $\Box$

{\bf Proof of Theorem 6} Let $x\in\rrr^n$ be a vector orthogonal to the $K$ principal
eigenvectors of $L$. Hence, by Proposition \ref{prop:perp} each
block $x_k$ of $x$ is orthogonal to $s_k$. In addition, for any pair
$k,l$ with $k\neq l$,
$[x_k'\;\;x_l']\left[\begin{array}{c}s_k'/\sqrt{\rho_k}\\s_l/\sqrt{\rho_l}]\\\end{array}\right]=0$,
    so $x$ is orthogonal to the Frobenius eigenvector of all the
    off-diagonal blocks $M_{kl}$.
We assume w.l.o.g. that $||x||=1$ and calculate
\beqa
|x'Lx|&=&|\sum_{k=1}^Kx_k'L_{kk}x_k+\sum_{k<l}[x_k' x_l']M_{kl}
\left[\begin{array}{l}x_k\\x_l\\\end{array}\right]\\
&\leq&\sum_{k=1}^Kr_{kk}|\lambda_2^{kk}|||x_k||^2
+\sum_{k<l}\sqrt{r_{kl}r_{lk}}|\lambda_2^{kl}|(||x_k||^2+||x_l||^2)\\
&\leq& c\sum_{k=1}^K||x_k||^2\left[r_{kk}+\sum_{k\neq l}\sqrt{r_{kl}r_{lk}}\right]\\
&\leq& c\max_{k}\left[r_{kk}+\sum_{k<l}\sqrt{r_{kl}r_{lk}}\right]
\eeqa
From this the result follows. \hfill $\Box$




\section{Matrix concentration results and the proof of theorem $3$} (main result)

\begin{proposition}[Modified theorem from Le and Vershynin \cite{le2015}]\label{prop:LaplacianConcentration}
(Concentration of the regularized Graph Laplacian) Let $\gdet, L, \Lhat$ have the usual meaning and let $\dmin$ be the minimum expected degree of $\gdet$, that is $\dmin = \min d_{1,\cdots, n}$, and denote $\dhatmin$ the analogous quantity for the observed degree of $\gdet$. Denote $d_{max} = \max_{ij}nS_{ij}$, $\gamma \ge 1$. $||.||$ denotes the spectral norm. If $(a)$ $\dhatmin \ge \log(n)$, $(b)$ $\dmin \ge \log(n)$, $(c)$ $\exists \varkappa>0$, $d  \le \varkappa\log n$, then with probability at least $1 - e^{-r}$,
\beq
||\Lhat - L|| \le \frac{\Psi \gamma^2}{\sqrt{\log n}}
\eeq
where $\Psi$ is a constant. 
\end{proposition}
{\bf Proof of Proposition \ref{prop:LaplacianConcentration}}
The proof mainly follows from \cite{le2015}. In the original theorem, they add equal weights to all entries of the similarity matrix to ensure the concentration of $L$. In this modified theorem, instead of adding weights we add the assumptions that $\dhatmin$ and $\dmin$ are bounded from below and prove that it will come to similar conclusion. \\
Denote $\tau = \log n$. In the step $1$ of their proof, we modify $E$ as $E:=\Lhat-L$, and get $E = M+T$, with $M = \Dhat^{-1/2}(A-S)\Dhat^{-1/2}$, and $T = \Dhat^{-1/2}S\Dhat^{-1/2} - D^{-1/2}SD^{-1/2}$. \\
In the step $2$ they give bound on $||M||$. We modify their $\Delta$ to be $\Delta_{ii} = 1$ if $\dhat_i \le 8\gamma d_{max}$ and $\Delta_{ii} = \dhat_i/\gamma$ otherwise. The rest of the proof in step $2$ still hold to this modification. We therefore obtain the bound for $M$ as $||M|| \le \frac{C_2\gamma^2}{\tau}(\sqrt{d_{max}} + \sqrt{\tau})$ with probability at least $1-2n^{-\gamma}$. $C_2$ is a constant. \\
We then follow the step $3$ of their proof and bound the spectral norm with the Hilbert-Schmidt norm. We get $||T|| \le ||T||_{HS} = \sum_{i,j=1}^n T_{ij}^2$, where $T_{ij} = S_{ij}[1/\sqrt{\delhat_{ij}} - 1/\sqrt{\delta_{ij}}]$ and $\delhat_{ij} = \dhat_i\dhat_j$ and $\delta_{ij}=d_id_j$. The rest of the proof in step $3$ can then be easily adapted to our modification and thus we obtain similar bound for $T$ as $||T||^2 \le \frac{C_6\gamma^4d^5}{\tau^6}$ with probability $1-e^{-2\gamma}$, $C_6$ a constant. Combining the results from step $2$ and step $3$ into the inequality D$||E|| \le ||M|| + ||T||$, we obtain $||E|| \le  \frac{C_7\gamma^2}{\sqrt{\tau}}[(d_{max}/\tau)^{5/2} + (d_{max}/r)^{1/2} + 1] \le \frac{\Psi \gamma^2}{\sqrt{\log n}}$ with high probability.

\begin{proposition} \label{prop:McSherry}
Let $L, \Lhat, Y, \Yhat$ have the usual meaning. Let $P_{\Lhat}$ denote the projection onto the span of $\Lhat$'s first $K$ left singular vectors, $\Lamhat$ and $\Lambda$ denote the diagonal matrices of the first $K$ eigenvalues of $\Lhat$ and $L$ accordingly. Then $P_{\Lhat}\Lhat  = \Yhat\Lamhat\Yhat^T$, and 
\beq
||P_{\Lhat}\Lhat - L||_F^2 = ||\Yhat\Lamhat\Yhat^T - Y\Lambda Y^T||_F^2 \le 8K||\Lhat-L||^2
\eeq
\end{proposition}
\yaliwan{We should probably put: $||.||$ denotes the spectral norm at the beginning of the paper and make the notations around the paper with $||.||$ clear}
\begin{proposition}[Davis-Kahan theorem] \label{prop:DavisKahan}
This is Davis-Kahan theorem\cite{stewart1990matrix} perturbation result. It puts results that relates the perturbation of $L$ to $Y$.\\
Let $S_0 \subset \rrr$ be an interval. Denote $Y_{S_0}$ as an orthonormal matrix whose column space is equal to the eigenspace of $L$ corresponding to the eigenvalues in $\lambda_{S_0}(L)$, where 
\beq
\lambda_{S_0}(L) = \{\{\lambda_1, \cdots, \lambda_n\} \cap S_0\}
\eeq
Denote by $\Yhat_{S_0}$ the analogous quantity for $P_{\Lhat}\Lhat$. Define the distance between $S_0$ and the spectrum of $L$ outside of $S_0$ as 
\beq
\Delta = \min\{|\lambda-s|;\lambda \text{ eigenvalue of L, }\lambda \notin S_0, s\in S_0\}
\eeq  
If $Y_{S_0}$ and $\Yhat_{S_0}$ are of the same dimension, then there is an orthogonal matrix $O$ that depends on $Y_{S_0}$ and $\Yhat_{S_0}$, such that
\beq
||Y_{S_0} - \Yhat_{S_0}||_F^2 \le \frac{2||P_{\Lhat} - L||_F^2}{\Delta^2}
\eeq
\end{proposition}
\yaliwan{notation $\lambda_{S_0}(L)$ should also be considered to put in the beginning, It is defined in the main theorem}
\begin{lemma} \label{lm:lm_2}
Assume $x_1, x_2, \cdots, x_n \in R^n$ form an orthonormal basis. then $||x_i - x_j||_2 = \sqrt{2}$, $\forall i, j \in \{ 1, \cdots, n\}$ 
\end{lemma}
{\bf Proof of Lemma \ref{lm:lm_2}}
define $X = (x_1, x_2,\cdots, x_n)$ Then $X^TX = I$. $\sqrt{2} = ||X^Tx_1 - X^Tx_2||_2 =(x_1 - x_2)^TXX^T(x_1 -x_2) = ||x_1 - x_2||_2$
\begin{lemma}\label{lm:lm_3}
 Assumption the H\pfm~model holds, the eigenvalues $\lambda_1, \cdots, \lambda_K$ of $R$ are the $K$ eigenvalues of $P$ that have the largest absolute values. The $i$th eigenvector of $P$ associated with these eigenvalues can be represented by the eigenvectors of $R$ as
\[
(\underbrace{u_{i1}, \cdots, u_{i1}}_{n_1}, \underbrace{u_{i2}, \cdots, u_{i2}}_{n_2}\cdots, \underbrace{u_{iK},\cdots,u_{iK}}_{u_K})
\]
$i \in \{1, \cdots, K\}$
\end{lemma}
{\bf Proof of Lemma \ref{lm:lm_3}}
Since $P$ is block stochastic and $R_{lk} = \sum_{j \in C_k}P_{ij}$, $i \in C_l, j\in C_k$. The first $K$ eigenvectors of $P$ are piecewise constant with respect to the clusters. Assume $x \in R^n$ is one of the first K eigenvectors of $P$ associated with $\lambda$.
\beq
Px = \lambda x
\eeq
Since $x$ is piecewise constant, we define a vector $u \in R^K$, $u_k = x_j$ if $\text{node } j \in C_k$.\\ Assume $\text{node } i \in C_l$.  
\beq
\lambda x_{i} = \sum_{j=1}^n[P_{ij}x_{j}] = \sum_{k=1}^K[\sum_{j \in C_k}P_{ij}x_{j}]
\eeq
\beq
 \text{that is, } \lambda u_{l} = \sum_{k=1}^K[R_{lk}u_{k}] 
\eeq
$i = 1, \cdots, n$.\\
Therefore $u$ is an eigenvector of $R$ associated with the eigenvalue $\lambda$.

\begin{lemma}\label{lm:lm_5}
Assume $d_i = \sum_{j=1}^nBernoulli(S_{ij})$, $S_{ij}\le1$, then 
\beq
P(|\sqrt{\hat{d}_i} - \sqrt{d_i}| \le \epsilon) \ge 1-2\exp[-\frac{\epsilon^2}{2 + \epsilon/\sqrt{d_i}}]
\eeq
\end{lemma}
{\bf Proof of Lemma \ref{lm:lm_5}}
Using Chernoff bound, we can get that,
\[
P[|\hat{d}_i-d_i| <\delta d_i] \ge 1- 2e^{-\frac{\delta^2 d_i}{2+\delta}}
\]
\[
P[|\sqrt{\hat{d}_i} - \sqrt{d_i}| <\delta\sqrt{d_i}] \ge P[|\sqrt{\hat{d}_i} - \sqrt{d_i}| <\frac{\delta d_i}{\sqrt{\dhat_i} + \sqrt{d_i}}] \ge 1- 2e^{-\frac{\delta^2d_i}{2+\delta}}
\]
Denote $\epsilon = \delta \sqrt{d_i}$, we can get 
\[
P(|\sqrt{\dhat_i} - \sqrt{d_i}| \le \epsilon) \ge 1-2\exp[-\frac{\epsilon^2}{2 + \epsilon/\sqrt{d_i}}]
\]

\yaliwan{we should probably add the condition that $\gmax$ should be bigger than 0, however I am not really sure how to add this condition} 

{\bf Proof of theorem $3$ from the paper}
Let $B, R, X, U, Y, V$ have the usual meaning. $\lambda_1, \cdots, \lambda_K$ are the eigenvalues of R. \\
In the H\pfm~model case, since $\lambda_{K+1} = \cdots=\lambda_n = 0$, the eigengap for $L$ between the first $K$ leading eigenvalues and the rest of the eigenvalues is $\lambda_K$. Denote $S_0 = (\frac{\lambda_K}{2},2) \subset \rrr$. \\
Using proposition \ref{prop:LaplacianConcentration}, when $n$ is sufficiently large, we obtain
\beq
|\lambda_K - \lamhat_K| \le ||L - \Lhat|| \le \frac{\Psi \gamma^2}{\sqrt{\log n}}  \le \lambda_K/10
\eeq
we therefore have $\lambda_{S_0}(\Lhat) = \{\lamhat_1,\cdots,\lamhat_K\}$.\\
\yaliwan{We can make $\max_i |\hat{\lamhat_i} - \lambda|$ smaller by modifying the constant of the right hand side of (c)}\\
\yaliwan{For general case, we will need to modify $S_0$}\\
Therefore $Y$ and $\Yhat$ are of the same dimension. Using proposition \ref{prop:DavisKahan}, we obtain,
\beq
\frac{1}{2}||\hat{Y} - Y\bigOO||_F^2 \le \frac{||P_{\Lhat}\Lhat-L||_F^2}{\Delta^2} 
\eeq
where $\bigOO$ is an orthonormal matrix. $\Delta = \lambda_K/2$.
Further apply proposition \ref{prop:McSherry}, we have the following inequality holds,
\beq
\frac{1}{2}||\Yhat - Y\bigOO||_F^2 \le \frac{4||P_{\Lhat}\Lhat-L||_F^2}{\lambda_K^2} \le \frac{32K||\Lhat-L||^2}{\lambda_K^2} \label{eq:DavisKahan}
\eeq
Now we take a closer look at $\frac{1}{2}||\Yhat - Y\bigOO||_F^2$ on the left hand side. 
\beq
\frac{1}{2}||\Yhat - Y\bigOO||_F^2 = \frac{1}{2}||D^{1/2}V-\Dhat^{1/2}\Vhat||_F^2 =  \frac{1}{2}||D^{1/2}(\Vhat-V) + (\Dhat^{1/2}-D^{1/2})\Vhat||^2_F
\eeq
Using \eqref{eq:DavisKahan}, we obtain, 
\beq
\underbrace{||D^{1/2}(\Vhat-V)||_F^2} _{(a)} \le \underbrace{||(\Dhat^{1/2}-D^{1/2})\Vhat||^2_F}_{(b)} +  \frac{64K||\Lhat-L||^2}{\lambda_K^2}
\eeq
\beq
(b) \le \max_i (d_i^{1/2} - \dhat_i^{1/2})^2||\Vhat||_F^2
\eeq
Using lemma \ref{lm:lm_5}, we can get,
\beq
P(|\sqrt{\dhat_i} - \sqrt{d_i}| \le \epps) \ge 1-2\exp[-\frac{\epsilon^2}{2 + \epsilon/\sqrt{d_i}}] \ge 1-2\exp[-\frac{\epsilon^2}{2 + \epsilon/\sqrt{\log n}}]  
\eeq
Meanwhile we have\\
\beq
||\Vhat||_F^2 = ||\Dhat^{-1/2}\Yhat||_F^2 \le \frac{1}{\dhat_{min}}||\Yhat||_F^2 = \frac{K}{\dhat_{min}} 
\eeq
Thus with probability at least $1-2\exp[-\frac{\epsilon^2}{2 + \epsilon/\sqrt{\log n}}]$
\beq
(b) \le \frac{K\epsilon^2}{\dhat_{min}}
\eeq


\beq
\frac{1}{2}||\hat{V}-V||_F^2\times \dmin \le(a)\le \frac{K\epsilon^2}{\dhat_{min}} + \frac{64K||\Lhat-L||^2}{\lambda_K^2}
\eeq
The above inequality gives us a bound for the perturbation of first $K$ eigenvectors of $P$.\\
Denote the $l_2$ norm of the perturbation for each row of $V$  is $e_{i}$, $i = 1,\cdots, n$. Denote the number of rows that have perturbation greater than $\frac{1}{2}\min_{i\not=j}||V(i,:)-V(j,:)||$ is $m$. Using proposition 4 from the paper, we have 
\beq
\frac{K\epsilon^2}{\dhat_{min}\dmin} + \frac{64K||\Lhat-L||^2}{\dmin\lambda_K^2} \ge \sum_{i=1}^n e_i^2 \ge \frac{m}{4\dtot}\gmax.
\eeq
Solving the above, and use proposition \ref{prop:LaplacianConcentration}. With probability at least $(1-2\exp[-\frac{\epsilon^2}{2 + \epsilon/\sqrt{\log n}}])(1 - e^{-r})$, we get, 
\beq
m \le \frac{C_0K\gamma^4\dtot}{\lambda_K^2\dmin\gmax\log n} + \frac{4K\dtot\epsilon^2}{\gmax\dhat_{min}\dmin}
\eeq
\beq
\err = m/n \le \frac{K\dtot}{n\dmin\gmax }\left[\frac{C_0\gamma^4}{\lambda_K^2\log n} + \frac{4\epsilon^2}{\dhat_{min}}\right]
\eeq


In the \pfm~case when $\lambda_K \not=0$, we can modify $S_0$ to be $S_0 = (\frac{\lambda_K+\lambda_{K+1}}{2}, 2)$, $\Delta = \sigma/2$, and then when n is sufficiently large $\lambda_{S_0}(\Lhat) = \{\lamhat_1,\cdots,\lamhat_K\}$ also holds. The rest of proof can be down similarly, and we obtain
\beq
\err \le \frac{K\dtot}{n\dmin\gmax }\left[\frac{C_0\gamma^4}{\sigma^2\log n} + \frac{4\epsilon^2}{\dhat_{min}}\right]
\eeq

\newpage
\bibliographystyle{plain}
\bibliography{kpfm}

\end{document}